\documentclass[letterpaper]{article} % DO NOT CHANGE THIS
\usepackage[preprint]{aaai2027} % DO NOT CHANGE THIS
\usepackage[hyphens]{url} % DO NOT CHANGE THIS
\usepackage{graphicx} % DO NOT CHANGE THIS
\usepackage{natbib} % DO NOT CHANGE THIS AND DO NOT ADD OPTIONS
\usepackage{caption} % DO NOT CHANGE THIS AND DO NOT ADD OPTIONS
\usepackage{amsmath}
\usepackage{amssymb}
\usepackage{booktabs}
\usepackage{array}
\usepackage{multirow}
\usepackage{colortbl}

\newcommand{\best}[1]{\textbf{#1}}

\definecolor{pptBlue}{HTML}{4472C4}
\definecolor{citeblue}{HTML}{2F5597}
\newcommand{\tablebest}[1]{\cellcolor{pptBlue!60}\textbf{#1}}
\newcommand{\tablesecond}[1]{\cellcolor{pptBlue!20}\textbf{#1}}

\title{Naka-GS: Naka-Guided Chroma Correction and Point-Prior Compaction for Low-Light 3D Gaussian Splatting}

\author{
    Runyu Zhu\textsuperscript{1},
    Sixun Dong\textsuperscript{1},
    Zhiqiang Zhang\textsuperscript{1},\\
    Qingxia Ye\textsuperscript{1},
    Rongjun Qin\textsuperscript{2},
    Zhihua Xu\textsuperscript{1}\thanks{Corresponding author.}
}

\affiliations{
    \textsuperscript{1}China University of Mining and Technology-Beijing, Beijing, China\\
    \textsuperscript{2}The Ohio State University, Columbus, OH, USA
}

\begin{document}

\maketitle

\begin{abstract}
Low-light inputs obscure scene content, distort color, and corrupt geometric priors, substantially degrading 3D Gaussian Splatting. Although image enhancement can recover visibility, it often introduces chromatic deviations and fine-scale artifacts that propagate through multi-view reconstruction. Moreover, dense point priors can further amplify these errors and increase optimization cost. To address the coupled challenges, we present Naka-GS, a low-light 3DGS framework that jointly improves observations and Gaussian initialization while retaining the standard 3DGS renderer and training objective. Our Naka-Guided Chroma Correction treats a fixed bionic response as a photometric anchor and learns bounded, frequency-decoupled luminance and chromatic corrections, preserving fine structures while suppressing residual color and boundary artifacts. The corrected views are reconstructed into dense scene priors, which a lightweight, non-learned Point-Prior Compaction Module refines through voxel pooling and distance-adaptive progressive pruning. On eight scene-disjoint RealX3D scenes, Naka-GS achieves 20.735~dB PSNR, 0.7408 SSIM, and 0.3340 LPIPS. PPM reduces initialization from 1.00M to 59.52K points and shortens average 3DGS optimization time by 37.7\%. 
\end{abstract}

% For anonymous review, any code/data link must itself be anonymous. Add an
% anonymous link only when the official AAAI-27 submission policy permits it.
% \begin{links}
%     \link{Code}{https://anonymous.4open.science/r/anonymous-repository}
% \end{links}
\begin{figure*}[t]
    \centering
    \includegraphics[width=\textwidth]{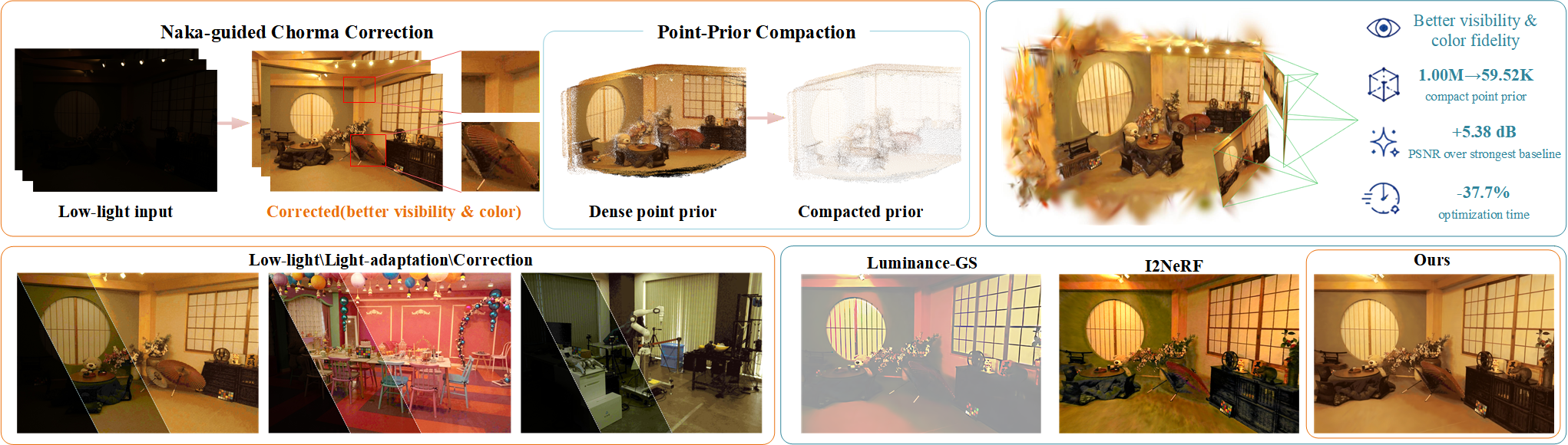}
    \caption{
   We present Naka-GS, a framework for low-light novel view synthesis, while restoring visibility, color fidelity, and reducing optimization time.
    }
    \label{fig:teaser}
\end{figure*}
\section{Introduction}
\label{sec:intro}

Reconstructing 3D scenes from low-light images remains challenging because insufficient illumination obscures visual content and introduces exposure imbalance, color distortion, and degraded local details. Although 3D Gaussian Splatting (3DGS) \cite{kerbl20233d} provides an efficient explicit representation for high-quality rendering, its performance deteriorates under severely underexposed observations. Image enhancement can recover visibility before reconstruction, but visibility-oriented processing may over-amplify bright regions, distort color, and introduce fine-scale texture artifacts. These residual errors can propagate through multi-view reconstruction and ultimately limit novel-view synthesis, making photometrically faithful observation restoration essential for low-light 3DGS.

The challenge does not end at image restoration. When the enhanced views are processed by feed-forward multi-view reconstruction models, dense point priors provide broad scene coverage but may also replicate residual texture artifacts and chromatic blotches across many locally redundant samples, amplifying their influence on subsequent 3DGS optimization. Simply reducing the number of predicted points is suboptimal, since scene-dependent density variations can lead to uneven spatial coverage and missing structures. Low-light 3DGS therefore requires not only reliable observation correction, but also adaptive compaction of dense geometric priors before Gaussian initialization, suppressing local redundancy while preserving sparsely covered regions.

To this end, we propose \textbf{Naka-GS}, a low-light 3DGS framework built around chroma correction and compact Gaussian initialization, as summarized in Fig.~\ref{fig:teaser}. Its first component, \textbf{Naka-Guided Chroma Correction (Naka-GCC)}, treats the analytical Naka response as a photometric anchor and learns bounded luminance and chromatic corrections, while preserving high-frequency structures from unnecessary modification. The corrected views are subsequently processed by a frozen feed-forward multi-view reconstruction model to obtain camera parameters and a dense colored point set. Dense feed-forward point priors provide broad scene coverage but also replicate residual image artifacts and introduce substantial local redundancy. A lightweight, non-learned \textbf{Point-Prior compaction Module (PPM)} therefore compacts the reconstructed prior through voxel pooling and density-adaptive pruning, preserving sparsely covered regions while reducing Gaussian initialization and optimization costs.

Experiments on RealX3D\cite{liu2025realx3d} scenes demonstrate the effectiveness of Naka-GS. Our method achieves the best performance among the evaluated baselines on all three metrics, with average scores of 20.735~dB PSNR, 0.7408 SSIM, and 0.3340 LPIPS. Meanwhile, PPM reduces the initialization point count from 1.00M to 59.52K and shortens the average 3DGS optimization time by 37.7\%, while largely preserving reconstruction quality.

Our contributions are summarized as follows:
\begin{itemize}
    \item We propose \textbf{Naka-GS}, a framework that combines Naka-guided image correction, feed-forward multi-view reconstruction, and compact Gaussian initialization for low-light 3DGS.

    \item We introduce a \textbf{Naka-GCC} model, which refines fixed Naka enhancement through bounded luminance and color correction together with frequency-decoupled reconstruction, reducing residual chromatic and local-structure artifacts.

    \item We develop a lightweight, non-learned \textbf{PPM} that refines dense point priors through voxel pooling and distance-adaptive progressive pruning, substantially improves Gaussian initialization quality and optimization efficiency with minimal overhead.
\end{itemize}

\section{Related Work}
\label{sec:related_work}

\subsection{Low-Light Novel View Synthesis}

Novel view synthesis has evolved from implicit radiance fields to explicit Gaussian-based representations. NeRF~\cite{mildenhall2021nerf} models scenes as continuous radiance fields, while 3D Gaussian Splatting (3DGS)~\cite{kerbl20233d} enables efficient real-time rendering with anisotropic Gaussian primitives. However, low-light inputs introduce poor visibility, sensor noise, exposure imbalance, and color distortion, which degrade both rendered appearance and geometry-related cues for reconstruction.

Existing low-light neural rendering methods incorporate illumination decomposition, degradation correction, camera-response modeling, or view-adaptive photometric adjustment into NeRF or 3DGS optimization\cite{wang2023llnerf,cui2024aleth,ye2024gaussian,you2024gaussian,cui2025luminance,zhou2025lita}. Other studies address RAW/HDR reconstruction, relighting, and robotic dark-scene perception\cite{jin2024lighting,li2024chaos,zhang2024darkgs,sun2025ll}. In contrast, we retain the standard 3DGS renderer and objective, focusing on observation correction and geometric-prior refinement.

\subsection{Bio-inspired photometric priors.}

Some low-light enhancement methods draw inspiration from biological visual mechanisms\cite{land1971lightness} for brightness adaptation and color constancy\cite{cai2023brain}. Retinex-based methods model an image as the composition of illumination and reflectance, providing an interpretable framework for low-light restoration\cite{land1971lightness,wei2018deep,cai2023retinexformer}. Related studies further extend this line through deep decomposition\cite{wei2018deep}, unrolled optimization\cite{liu2021retinex,wu2022uretinex}, and prior-guided restoration\cite{wang2024zero}. Different from these methods, our work does not aim to design a biologically grounded enhancement architecture. Instead, inspired by Naka-Rushton function\cite{naka1966s}, we utilize it as a photometric prior, and then try to efficiently correct the dominant photometric errors observed after Naka-based enhancement.

\subsection{Feed-Forward Geometry Priors and Gaussian Initialization}

The quality of geometric priors strongly affects 3DGS initialization. Classical pipelines rely on SfM or MVS to estimate camera poses and point clouds before Gaussian optimization. Recently, feed-forward geometry models have provided dense multi-view priors without conventional reconstruction pipelines. DUSt3R~\cite{wang2024dust3r} predicts dense pointmaps for unconstrained stereo reconstruction, MASt3R~\cite{leroy2024mast3r} grounds image matching in 3D, and VGGT~\cite{wang2025vggt} jointly predicts camera parameters, point maps, depth maps, and tracks in a feed-forward transformer.

Feed-forward models such as DUSt3R, MASt3R, and VGGT provide dense multi-view geometry priors without conventional SfM/MVS. However, residual texture artifacts and chromatic blotches in enhanced low-light views can be replicated across dense predictions, whereas globally limiting the prediction count risks scene-dependent coverage loss. We therefore compact the reconstructed prior adaptively, suppressing local redundancy while preserving sparsely sampled regions.
\section{Method}
\label{sec:method}

\begin{figure*}[t!]
    \centering
    \includegraphics[width=0.95\textwidth,trim={119bp 190bp 103bp 166bp},clip]{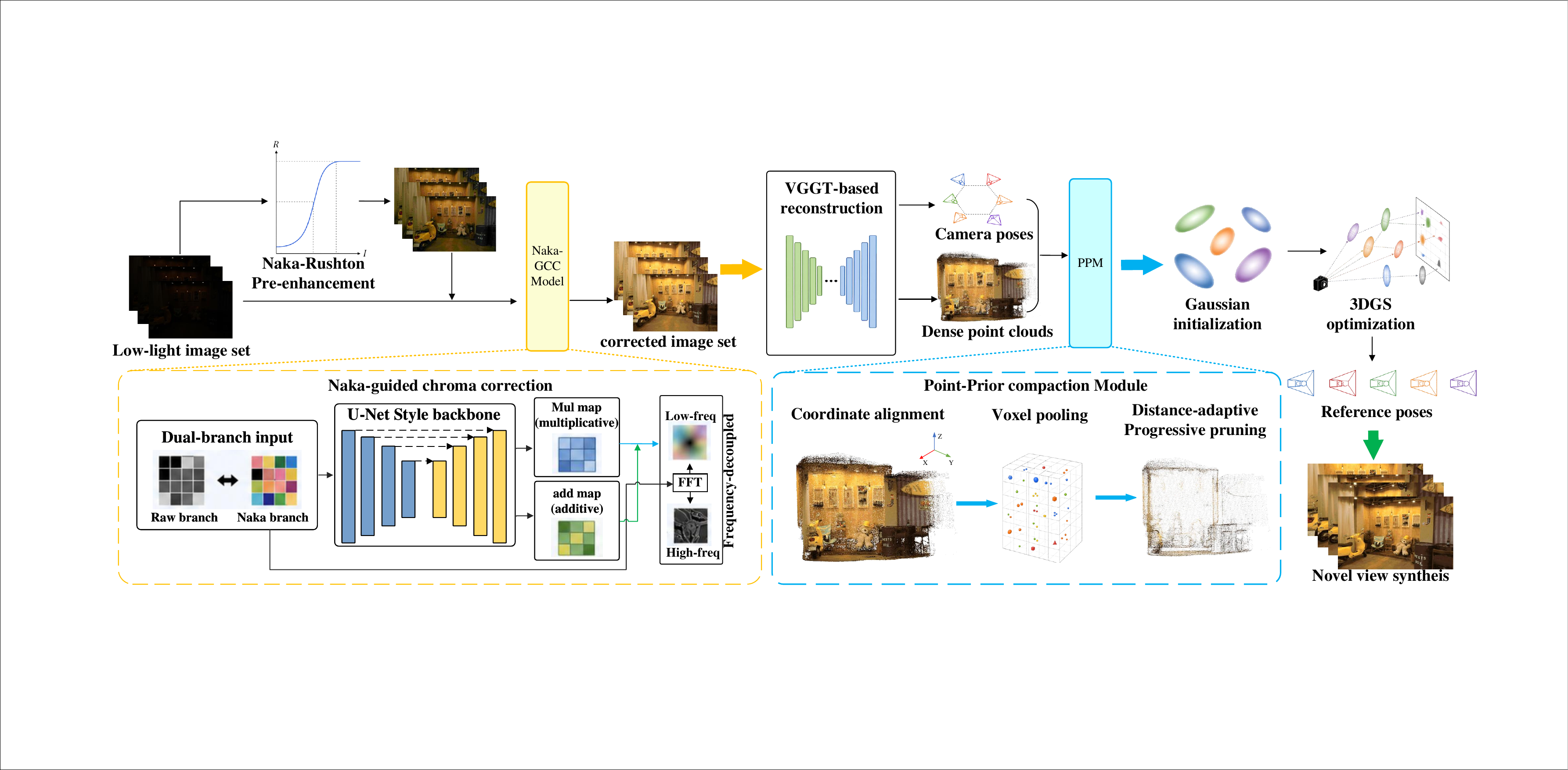}
    \vspace{-0.8em}
    \caption{Overview of the proposed Naka-GS pipeline. The pipeline consists of three parts: Naka-guided Chroma-Correction model, frozen VGGT model for multi-view reconstruction, and Gaussian Splatting with Point-Prior compaction Module(PPM).}
    \label{fig:overview}
    \vspace{-0.8em}
\end{figure*}

\subsection{Overview}

Given a set of low-light views
$\mathcal{I}^{\mathrm{low}}=\{I_i^{\mathrm{low}}\}_{i=1}^{N}$,
our goal is to recover a 3D Gaussian representation that supports
photometrically faithful novel-view synthesis under severe illumination
degradation. To achieve this goal, we propose Naka-GS, a low-light 3DGS framework that addresses two sources of degradation at observation and initialization levels, while leaving the Gaussian render itself unchanged. As illustrated in Fig.~\ref{fig:overview}, it first utilizes a fixed photometric anchor and learns bounded residual corrections within Naka-Guided Chroma Correction. Then the corrected images are fed into a frozen multi-view reconstruction model to predict a dense colored point cloud set that serves as the geometric prior for the scene. Next, to mitigate the effects of dark degradation amplified by dense prior, the Point-Prior compaction Module (PPM) prunes this dense prior into a compact, spatially representative initialization. Finally, a standard 3DGS optimizing pipeline is applied to get faithful NVS images.

\subsection{Naka-Guided Chroma Correction}

A fixed nonlinear response can reveal nearly invisible content
in the input, but it cannot adapt to scene-dependent color shifts,
highlight compression, or boundary artifacts. A lightweight correction network predicts identity-centered photometric adjustments, and these adjustments are
restricted to the low-frequency component so that fine image structures
are not unnecessarily rewritten.

\paragraph{Naka--Rushton photometric anchor.}
For an input intensity $I\in[0,1]$, the Naka--Rushton\cite{naka1966s} response is
\begin{equation}
R(I)=\frac{I^n}{I^n+\sigma^n},
\label{eq:naka_response}
\end{equation}
where $\sigma$ is the half-saturation constant and $n$ controls the
response slope. We fix $n=1.0$ and $\sigma=0.05$ in all experiments.
The same transform is applied independently to the three RGB channels
of every view, followed by the same fixed mapping to $[0,1]$. Sharing
the response parameters and preprocessing rule across views avoids
introducing view-specific photometric adaptation at this stage.

Let $I^{\mathrm{naka}}$ denote the transformed image. The response
expands dark intensities and substantially improves visibility, but its
channel-wise saturation can still cause color deviations. It may also amplify weak noise or inconsistent local contrast around textures and object boundaries. Naka-GCC is designed to correct these residual errors while retaining the visibility gain
provided by the analytical transform.

\paragraph{Dual-branch input representation.}
For each view, we first form the residual between the transformed and
raw observations,
\begin{equation}
\Delta I = I^{\mathrm{naka}}-I^{\mathrm{low}}.
\label{eq:naka_residual}
\end{equation}
The correction network receives both absolute photometric measurements
and their standardized counterparts:
\begin{equation}
\begin{aligned}
X=\operatorname{Concat}\big(
&I^{\mathrm{low}}, I^{\mathrm{naka}}, \Delta I,\\
&\mathcal{S}(I^{\mathrm{low}}),
\mathcal{S}(I^{\mathrm{naka}}),
\mathcal{S}(\Delta I)
\big),
\end{aligned}
\label{eq:dual_input}
\end{equation}
where
\begin{equation}
\mathcal{S}(x)=\frac{x-\mu_x}{\sigma_x+\varepsilon}
\label{eq:standardization}
\end{equation}
denotes per-tensor standardization. The unnormalized tensors retain
absolute intensity and color information, which is necessary for
photometric reconstruction. Their standardized counterparts reduce the
dominance of the global exposure scale and emphasize relative contrast.
Together, the resulting 18-channel representation allows the network
to distinguish global brightening from spatially localized chromatic
and structural errors.

\paragraph{Identity-centered bounded correction.}
We build a lightweight U-Net-style encoder--decoder with skip
connections, residual blocks, and channel attention. The network predicts a single-channel
multiplicative field for luminance modulation and a three-channel
additive field for color correction. The unconstrained outputs
$F_{\mathrm{mul}}(X)$ and $F_{\mathrm{add}}(X)$ are mapped to
\begin{equation}
M_{\mathrm{mul}}
=
1+\alpha_{\mathrm{mul}}
\tanh\!\left(F_{\mathrm{mul}}(X)\right),
\label{eq:mul_map}
\end{equation}
\begin{equation}
M_{\mathrm{add}}
=
\alpha_{\mathrm{add}}
\tanh\!\left(F_{\mathrm{add}}(X)\right).
\label{eq:add_map}
\end{equation}
We set $\alpha_{\mathrm{mul}}=0.6$ and
$\alpha_{\mathrm{add}}=0.25$. The offset in
Eq.~\eqref{eq:mul_map} centers multiplicative correction at the
identity, giving
$M_{\mathrm{mul}}\in[1-\alpha_{\mathrm{mul}},
1+\alpha_{\mathrm{mul}}]$, while
$M_{\mathrm{add}}\in[-\alpha_{\mathrm{add}},
\alpha_{\mathrm{add}}]$. This parameterization constrains the network
to refine the Naka-based estimate instead of replacing it with an
unrestricted image prediction. The shared multiplicative field adjusts
low-frequency luminance consistently across channels, whereas the
additive RGB field models channel-specific residual color errors.

\paragraph{Frequency-decoupled reconstruction.}
Applying learned correction directly to the full-resolution image can
over-smooth textures or alter high-frequency structures that support
multi-view correspondence. We therefore decompose
$I^{\mathrm{naka}}$ into low- and high-frequency components:
\begin{equation}
I^{\mathrm{lf}}=G(I^{\mathrm{naka}}),
\qquad
I^{\mathrm{hf}}=I^{\mathrm{naka}}-I^{\mathrm{lf}},
\label{eq:frequency_decomposition}
\end{equation}
where $G(\cdot)$ is a Gaussian filter with kernel size $5$ and standard
deviation $1.0$. The learned maps operate only on the low-frequency
component,
\begin{equation}
I^{\mathrm{base}}
=
I^{\mathrm{lf}}\odot M_{\mathrm{mul}}+M_{\mathrm{add}},
\label{eq:base_correction}
\end{equation}
and the final corrected image is reconstructed as
\begin{equation}
\hat{I}
=
\operatorname{clip}
\left(I^{\mathrm{base}}+I^{\mathrm{hf}},0,1\right).
\label{eq:corrected_image}
\end{equation}
This separation assigns smooth illumination and chromatic correction
to the learned branch while passing the original high-frequency
residual directly to the output. It thereby reduces unnecessary
modification of edges and fine textures.

\subsection{Learning Objective}

Naka-GCC is trained with complementary photometric, structural,
perceptual, and regularization terms:
\begin{equation}
\begin{aligned}
\mathcal{L}={}&
\lambda_{\mathrm{rgb}}\mathcal{L}_{\mathrm{rgb}}
+\lambda_{\mathrm{mse}}\mathcal{L}_{\mathrm{mse}}
+\lambda_{\mathrm{chroma}}\mathcal{L}_{\mathrm{chroma}}\\
&+\lambda_{\mathrm{ssim}}\mathcal{L}_{\mathrm{ssim}}
+\lambda_{\mathrm{edge}}\mathcal{L}_{\mathrm{edge}}
+\lambda_{\mathrm{feat}}\mathcal{L}_{\mathrm{feat}}\\
&+\lambda_{\mathrm{reg}}\mathcal{L}_{\mathrm{reg}}
+\lambda_{\mathrm{gray}}\mathcal{L}_{\mathrm{gray}} .
\end{aligned}
\label{eq:total_loss}
\end{equation}

The RGB reconstruction loss combines Charbonnier and $\ell_1$
penalties:
\begin{equation}
\mathcal{L}_{\mathrm{rgb}}
=
\mathcal{L}_{\mathrm{char}}
+
0.5\mathcal{L}_{1},
\end{equation}
where
\begin{equation}
\mathcal{L}_{\mathrm{char}}
=
\frac{1}{N}\sum_{q=1}^{N}
\sqrt{
\left(\widehat{I}^{q}-I_{\mathrm{gt}}^{q}\right)^2
+\epsilon_{\mathrm{char}}^2
},
\quad
\epsilon_{\mathrm{char}}=10^{-3},
\end{equation}
and
$\mathcal{L}_{1}=\frac{1}{N}\sum_{q=1}^{N}
|\widehat{I}^{q}-I_{\mathrm{gt}}^{q}|$.
We further employ the mean squared error
$\mathcal{L}_{\mathrm{mse}}
=\frac{1}{N}\sum_{q=1}^{N}
(\widehat{I}^{q}-I_{\mathrm{gt}}^{q})^{2}$,
which more strongly penalizes large reconstruction residuals.

The chroma loss constrains luminance and chrominance consistency in
YCbCr space, while $\mathcal{L}_{\mathrm{ssim}}$ encourages structural
similarity. For edge preservation, $\mathcal{L}_{\mathrm{edge}}$
computes Sobel gradient magnitudes independently on the RGB channels
and penalizes their $\ell_1$ difference:
\begin{equation}
\begin{aligned}
\mathcal{L}_{\mathrm{edge}}
&=
\left\|
E(\widehat{I})-E(I_{\mathrm{gt}})
\right\|_1,\\
E(I)
&=
\sqrt{G_x(I)^2+G_y(I)^2+10^{-6}} .
\end{aligned}
\end{equation}
where $G_x$ and $G_y$ denote horizontal and vertical Sobel responses.
This term is separate from $\mathcal{L}_{\mathrm{gray}}$, which applies
an additional gray-scale edge-aware reconstruction constraint.

The perceptual loss is computed using an ImageNet-pretrained VGG-19
network with frozen parameters. Let
$\mathcal{V}=\{\mathrm{relu1\_2},\mathrm{relu2\_2},
\mathrm{relu3\_4},\mathrm{relu4\_4}\}$ denote the selected VGG-19
layers. We use equally weighted $\ell_1$ distances over these features:
\begin{equation}
\mathcal{L}_{\mathrm{feat}}
=
\frac{1}{4}
\sum_{l\in\mathcal{V}}
\left\|
\phi_l(\widehat{I})-\phi_l(I_{\mathrm{gt}})
\right\|_1 ,
\end{equation}
where the images are normalized with ImageNet mean and standard
deviation before being fed into VGG-19.

The regularization term constrains both the magnitude and spatial
variation of the predicted multiplicative and additive correction maps:
\begin{equation}
\begin{aligned}
\mathcal{L}_{\mathrm{reg}}
={}&
\|M_{\mathrm{mul}}-1\|_1
+\|M_{\mathrm{add}}\|_1 \\
&+0.5\left(
\mathcal{L}^{\mathrm{mul}}_{\mathrm{smooth}}
+
\mathcal{L}^{\mathrm{add}}_{\mathrm{smooth}}
\right) \\
&+0.1\,\mathrm{ReLU}
\left(
\|\widehat{I}-I_{\mathrm{gt}}\|_1
-
\|I_{\mathrm{Naka}}-I_{\mathrm{gt}}\|_1
\right).
\end{aligned}
\end{equation}
Here, the first two terms encourage the correction maps to stay close
to the identity transformation, the smoothness terms are first-order
horizontal and vertical total-variation penalties, and the last term
discourages the learned correction from becoming worse than the
Naka-processed input.

To emphasize reconstruction accuracy around structural boundaries, we
derive a spatial weighting mask from the ground-truth image. We first
convert the ground-truth image to grayscale as
\begin{equation}
Y_{\mathrm{gt}}
=
0.299 I_{\mathrm{gt}}^{R}
+0.587 I_{\mathrm{gt}}^{G}
+0.114 I_{\mathrm{gt}}^{B}.
\end{equation}
Let
\begin{equation}
A = \left|\Delta_{3}Y_{\mathrm{gt}}\right|,
\qquad
M_{\mathrm{gray}}
=
\sqrt{
\frac{A}
{\max_{u,v} A(u,v)+\epsilon_{\mathrm{gray}}}
},
\end{equation}
where $\Delta_{3}$ denotes a $3\times3$ Laplacian operator and
$\epsilon_{\mathrm{gray}}=10^{-8}$. The normalization is performed
independently for each image, while the square-root operation compresses
extreme responses and retains weaker structural boundaries. The
gray-edge loss is then defined as
\begin{equation}
\mathcal{L}_{\mathrm{gray}}
=
\frac{1}{3BHW}
\sum_{b,c,u,v}
M_{\mathrm{gray}}^{b}(u,v)
\left|
\hat{I}^{b,c}(u,v)-I_{\mathrm{gt}}^{b,c}(u,v)
\right|,
\end{equation}
where the single-channel mask is shared across the three RGB channels.

In all experiments, we set
$\lambda_{\mathrm{rgb}}=1.0$,
$\lambda_{\mathrm{mse}}=0.05$,
$\lambda_{\mathrm{chroma}}=0.5$,
$\lambda_{\mathrm{ssim}}=0.3$,
$\lambda_{\mathrm{edge}}=0.2$,
$\lambda_{\mathrm{feat}}=0.15$,
$\lambda_{\mathrm{reg}}=0.02$, and
$\lambda_{\mathrm{gray}}=1.0$.
During training, image pairs are first randomly rescaled with
$s\sim\mathcal{U}(0.7,1.4)$, then randomly cropped into $(256\times256)$ patches. During validation and inference, Naka-GCC is fully convolutional and directly processes the original full-resolution images without resizing or cropping. 

\subsection{Point-Prior compaction Module for Gaussian Initialization}

\begin{figure*}[t]
    \centering
    \includegraphics[width=0.95\textwidth,trim={175bp 290bp 140bp 275bp},clip]{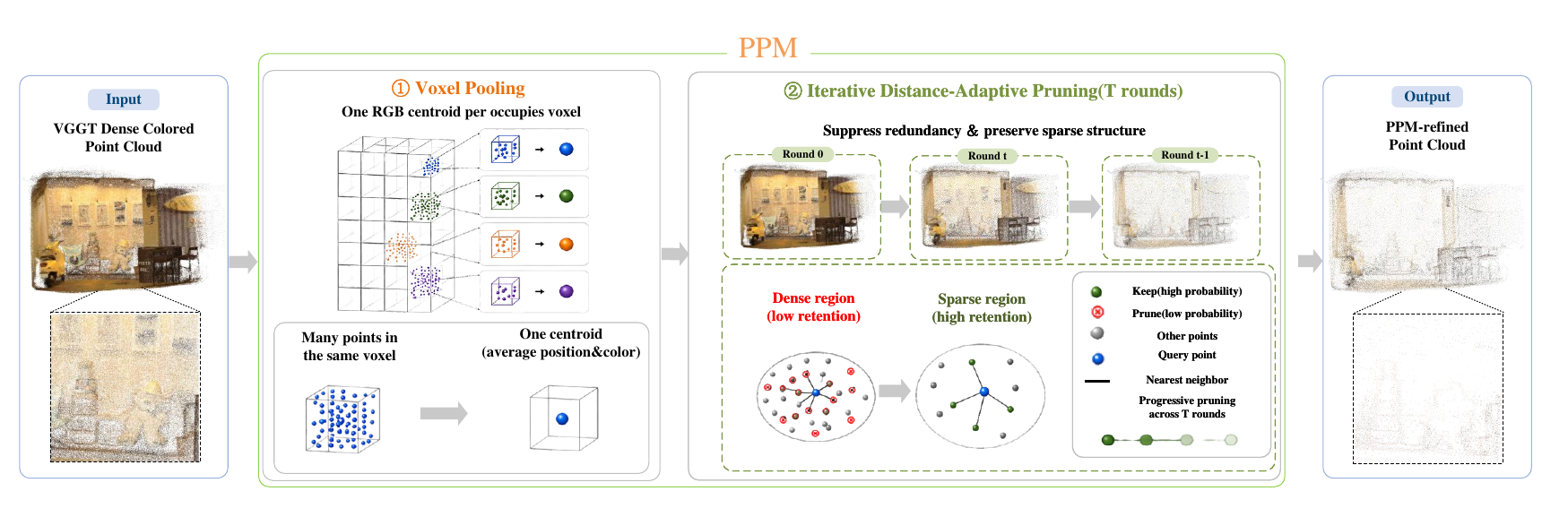}
    \caption{Overview of PPM, which voxel pools, and progressively prunes the reconstructed point cloud before Gaussian initialization.}
    \label{fig:architecture_ppm}
\end{figure*}

The corrected views are passed to a feed-forward multi-view geometry
model, which predicts camera parameters and a dense colored point set
$\mathcal{P}_0$. Dense prediction provides broad scene coverage, but it also introduces many near-duplicate samples in locally over-observed regions, which may expand the adverse effect brought by remaining fine-scale texture artifacts and chromatic blotches. Moreover, initializing one Gaussian from every predicted point therefore increases optimization cost without proportionally increasing spatial coverage. PPM converts $\mathcal{P}_0$ into a compact point set by preserving spatially isolated samples and aggregating or thinning points in locally dense regions.

\paragraph{Coordinate alignment.}
When ground-truth camera poses are provided, we estimate a global
similarity transform between the reconstructed and target camera
centers and apply it to the predicted point cloud so that the point
prior is expressed in the target 3DGS coordinate frame.

\paragraph{Voxel Pooling.}
We first remove obvious local redundancy by partitioning the aligned
point set into regular voxels. For each occupied voxel, all contained
points are replaced by one representative whose coordinate and RGB
value are the corresponding within-voxel means. The resulting set $\mathcal{P}_v$ provides a compact
candidate prior for subsequent density-adaptive thinning.

\paragraph{Distance-adaptive iterative thinning.}
Voxel pooling removes exact local crowding, but point density may
remain highly non-uniform across the scene. Let $a_k$ be a candidate
point in the current set and let $d_{\mathrm{nn}}(a_k)$ denote its
nearest-neighbor distance. At iteration $t$, we retain $a_k$ with
probability
\begin{equation}
P_{\mathrm{keep}}(a_k)
=
\min\left(
1,
\frac{d_{\mathrm{nn}}(a_k)}
{\tau^{(t)}+\varepsilon}
\right),
\label{eq:keep_probability}
\end{equation}
where $\tau^{(t)}$ controls the pruning scale. Points in crowded
regions have small nearest-neighbor distances and are therefore
sampled less frequently, whereas spatially isolated points are retained
with high probability. This realizes density-aware thinning rather
than uniform subsampling, which would remove points independently of
their geometric coverage.

After each accepted iteration, the threshold is updated as
\begin{equation}
\tau^{(t+1)}
=
\tau^{(t)}
\exp\left(
\beta\frac{M_t}{M_0}
\right),
\label{eq:threshold_update}
\end{equation}
where $M_0$ is the size of the initial candidate set,
$M_t$ is the number of retained points after iteration $t$, and
$\beta$ controls the update rate. Increasing $\tau^{(t)}$ gradually
tightens the retention probability, yielding progressive rather than
single-step thinning. To avoid collapse in weakly covered regions, we
enforce a minimum point count. If an iteration falls below this limit,
the update is rejected, the previous valid point set is restored, and
the procedure terminates.

\paragraph{3DGS initialization.}
The final compact point set $\widetilde{\mathcal{P}}$ is passed to the
standard 3DGS initializer. Its coordinates initialize Gaussian centers,
and its associated colors are used by the standard appearance
initialization. PPM is applied only once and introduces no additional
learned parameters or rendering-time computation. All remaining
Gaussian attributes, the rendering loss, densification, and the
optimization schedule follow the unmodified 3DGS pipeline.

\section{Experiments}
\label{subsec:experiment}

\begin{table*}[!t]
\centering
\scriptsize
\setlength{\tabcolsep}{2.8pt}
\renewcommand{\arraystretch}{0.96}
\caption{Quantitative comparison on eight unseen RealX3D scenes. Darker/lighter blue marks the best/second-best results; \textit{BlueHawaii} is excluded due to training overlap.}
\label{tab:quant_main}
\vspace{-0.45em}
\resizebox{0.94\textwidth}{!}{%
\begin{tabular}{@{}>{\raggedright\arraybackslash}p{2.55cm}|c|c|c|c|c|c|c|c|c|c@{}}
\specialrule{1.1pt}{0pt}{0pt}
\textbf{Methods} & \textbf{Metrics} & Chocolate & Cupcake & GearWorks & Laboratory & MilkCookie & Popcorn & Sculpture & Ujikintoki & \textbf{Avg.} \\
\midrule
\multirow{3}{2.55cm}{\shortstack[l]{3DGS\\ \textcolor{citeblue}{\citeauthor{kerbl20233d}~(\citeyear{kerbl20233d})}}}
& PSNR$\uparrow$ & 7.89 & 4.42 & 6.90 & 7.59 & 6.15 & 7.46 & 5.85 & 6.23 & 6.56 \\
& SSIM$\uparrow$ & 0.084 & 0.064 & 0.081 & 0.041 & 0.052 & 0.062 & 0.018 & 0.062 & 0.058 \\
& LPIPS$\downarrow$ & 0.645 & 0.625 & 0.658 & 0.610 & 0.653 & 0.650 & 0.733 & 0.683 & 0.657 \\
\midrule
\multirow{3}{2.55cm}{\shortstack[l]{AlethNeRF\\ \textcolor{citeblue}{\citeauthor{cui2024aleth}~(\citeyear{cui2024aleth})}}}
& PSNR$\uparrow$ & 11.53 & 13.49 & 8.82 & 14.14 & 13.49 & 13.25 & 12.22 & 14.12 & 12.63 \\
& SSIM$\uparrow$ & 0.406 & 0.552 & 0.217 & 0.484 & 0.567 & 0.452 & 0.208 & 0.544 & 0.429 \\
& LPIPS$\downarrow$ & 0.692 & 0.612 & 0.671 & 0.696 & 0.725 & 0.688 & 0.891 & 0.646 & 0.703 \\
\midrule
\multirow{3}{2.55cm}{\shortstack[l]{Luminance-GS\\ \textcolor{citeblue}{\tiny Cui, Chu,}\\ \textcolor{citeblue}{\tiny and Harada (2025)}}}
& PSNR$\uparrow$ & 7.33 & \tablesecond{14.82} & 8.09 & 8.88 & 9.67 & 11.33 & 9.79 & 9.36 & 9.91 \\
& SSIM$\uparrow$ & 0.362 & 0.536 & 0.411 & 0.372 & 0.532 & 0.462 & 0.293 & 0.463 & 0.429 \\
& LPIPS$\downarrow$ & 0.779 & 0.534 & 0.737 & 0.614 & 0.766 & 0.625 & 0.655 & 0.882 & 0.699 \\
\midrule
\multirow{3}{2.55cm}{\shortstack[l]{LITA-GS\\ \textcolor{citeblue}{\tiny Zhou, Dong,}\\ \textcolor{citeblue}{\tiny and Chen (2025)}}}
& PSNR$\uparrow$ & 17.94 & 13.07 & 10.90 & \tablesecond{17.56} & 12.74 & \tablesecond{18.97} & \tablesecond{12.84} & 18.82 & \tablesecond{15.36} \\
& SSIM$\uparrow$ & 0.541 & \tablesecond{0.643} & 0.344 & 0.597 & 0.573 & \tablesecond{0.557} & \tablesecond{0.343} & 0.656 & 0.532 \\
& LPIPS$\downarrow$ & \tablesecond{0.557} & \tablesecond{0.326} & \tablesecond{0.521} & \tablesecond{0.435} & \tablesecond{0.478} & \tablesecond{0.448} & \tablesecond{0.588} & 0.497 & \tablesecond{0.481} \\
\midrule
\multirow{3}{2.55cm}{\shortstack[l]{I2-NeRF\\ \textcolor{citeblue}{\citeauthor{liu2025i2nerf}~(\citeyear{liu2025i2nerf})}}}
& PSNR$\uparrow$ & \tablesecond{19.77} & 12.68 & \tablesecond{12.22} & 16.34 & \tablesecond{14.78} & 17.08 & 9.64 & \tablesecond{19.02} & 15.19 \\
& SSIM$\uparrow$ & \tablesecond{0.571} & 0.608 & \tablesecond{0.511} & \tablesecond{0.630} & \tablesecond{0.649} & 0.543 & 0.277 & \tablesecond{0.666} & \tablesecond{0.557} \\
& LPIPS$\downarrow$ & 0.561 & 0.493 & 0.543 & 0.486 & 0.544 & 0.488 & 0.648 & \tablesecond{0.474} & 0.530 \\
\midrule
\multirow{3}{2.55cm}{\shortstack[l]{NAKA-GS\\ \textcolor{citeblue}{(ours)}}}
& PSNR$\uparrow$ & \tablebest{20.63} & \tablebest{21.84} & \tablebest{17.94} & \tablebest{22.58} & \tablebest{21.11} & \tablebest{22.05} & \tablebest{14.57} & \tablebest{25.16} & \tablebest{20.735} \\
& SSIM$\uparrow$ & \tablebest{0.634} & \tablebest{0.878} & \tablebest{0.740} & \tablebest{0.813} & \tablebest{0.840} & \tablebest{0.733} & \tablebest{0.469} & \tablebest{0.819} & \tablebest{0.7408} \\
& LPIPS$\downarrow$ & \tablebest{0.405} & \tablebest{0.189} & \tablebest{0.383} & \tablebest{0.316} & \tablebest{0.290} & \tablebest{0.278} & \tablebest{0.469} & \tablebest{0.342} & \tablebest{0.3340} \\
\specialrule{1.1pt}{0.45ex}{0pt}
\end{tabular}%
}
\end{table*}

\subsection{Experimental Setup}

\paragraph{Dataset and evaluation protocol.}

The RealX3D~\cite{liu2025realx3d} benchmark, which provides physically captured low-light scenes for evaluating restoration and reconstruction, contains nine real-world low-light scenes. The Naka-GCC model is trained on five subsets of the LOM dataset\cite{cui2024aleth}, which offer paired low-light and normal-light multi-view images, together with the provided training images from \textit{BlueHawaii}. Since \textit{BlueHawaii} is not scene-disjoint from the training data, we exclude it from the primary benchmark and evaluate the proposed method on the remaining eight scenes: \textit{Chocolate}, \textit{Cupcake}, \textit{GearWorks}, \textit{Laboratory}, \textit{MilkCookie}, \textit{Popcorn}, \textit{Sculpture}, and \textit{Ujikintoki}. These scenes were not used to train Naka-GCC. We report PSNR, SSIM, and LPIPS following the standard novel view synthesis protocol. Higher PSNR and SSIM and lower LPIPS indicate better reconstruction quality.

\paragraph{Baselines.}
We compare Naka-GS with representative low-light reconstruction and novel view synthesis methods, including 3DGS~\cite{kerbl20233d}, AlethNeRF~\cite{cui2024aleth}, Luminance-GS~\cite{cui2025luminance}, LITA-GS~\cite{zhou2025lita}, and I2-NeRF~\cite{liu2025i2nerf}. These baselines cover both NeRF-based and 3DGS-based low-light reconstruction paradigms. All methods are evaluated under the same scene-level protocol.

\paragraph{Implementation details.}
Naka-GCC is trained for 200 epochs using AdamW with an initial learning rate of $2\times10^{-4}$. PPM uses a voxel size of 0.01, an initial threshold of 0.005, an update factor $\beta=0.01$, six thinning iterations, and a minimum-retention constraint. The subsequent 3DGS optimization runs for 8,000 steps on a single RTX A6000 GPU. Detailed architecture, augmentation, and optimization details are provided in the supplementary material.

\paragraph{Model complexity.}
Naka-GCC contains 5.44M trainable parameters. The network requires 56.06 GFLOPs for a $(256\times256)$ input. During inference, however, Naka-GCC is applied directly to the original evaluation images without cropping. Under this full-resolution inference setting, it achieves an average network-forward latency of 276.95+/-17.53 ms per image with batch size one on an NVIDIA RTX A6000. Detailed complexity, runtime, and memory statistics are provided in the supplementary material.

\subsection{Quantitative Comparison}

Table~\ref{tab:quant_main} compares Naka-GS with representative low-light NeRF and 3DGS methods on the eight scene-disjoint RealX3D test scenes. Naka-GS ranks first on all three metrics for every evaluated scene and achieves average scores of 20.735~dB PSNR, 0.7408 SSIM, and 0.3340 LPIPS. Relative to the strongest competing method for each metric, Naka-GS improves PSNR by 5.38~dB over LITA-GS, improves SSIM by 0.184 over I2-NeRF, and reduces LPIPS by 0.147 compared with LITA-GS. The consistent improvements across all eight test scenes demonstrate that the proposed framework transfers effectively beyond the \textit{BlueHawaii} scene used during Naka-GCC training.

\subsection{Qualitative Comparison}

\begin{figure*}[t]
    \centering
    \includegraphics[width=0.96\textwidth]{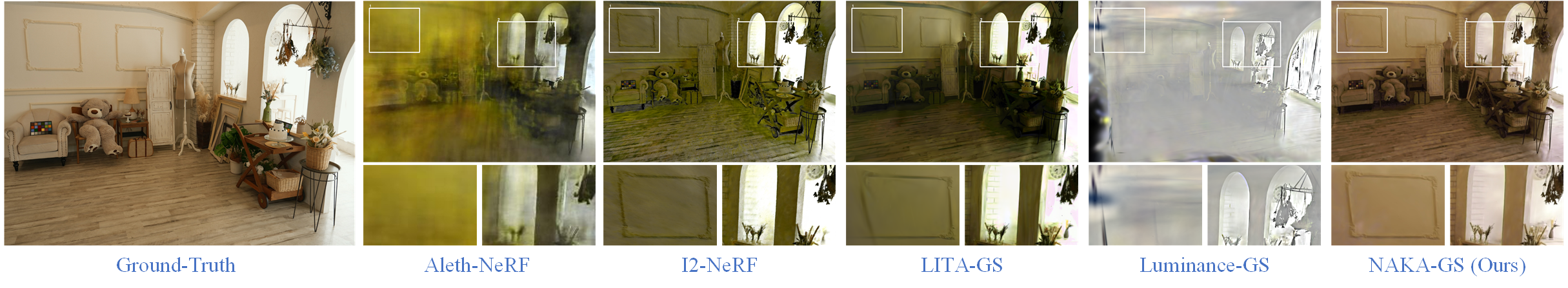}\\[-0.35em]
    \includegraphics[width=0.96\textwidth]{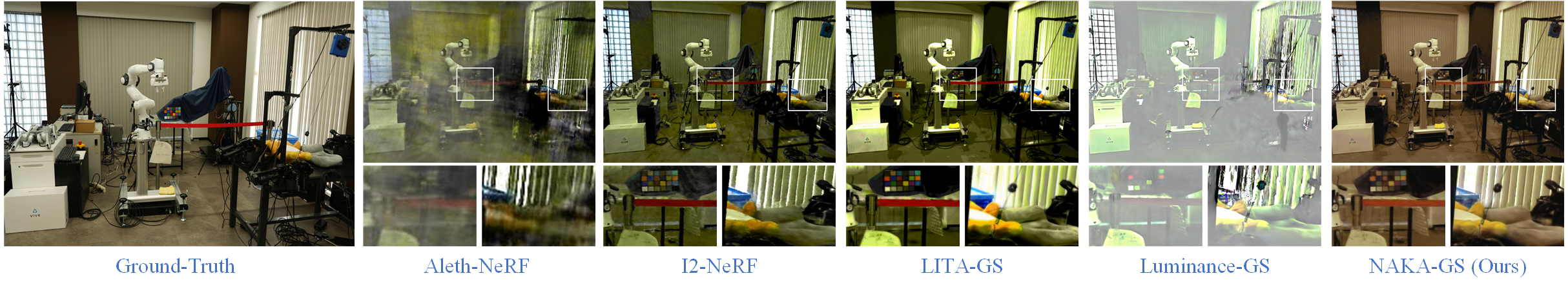}\\[-0.35em]
    \includegraphics[width=0.96\textwidth]{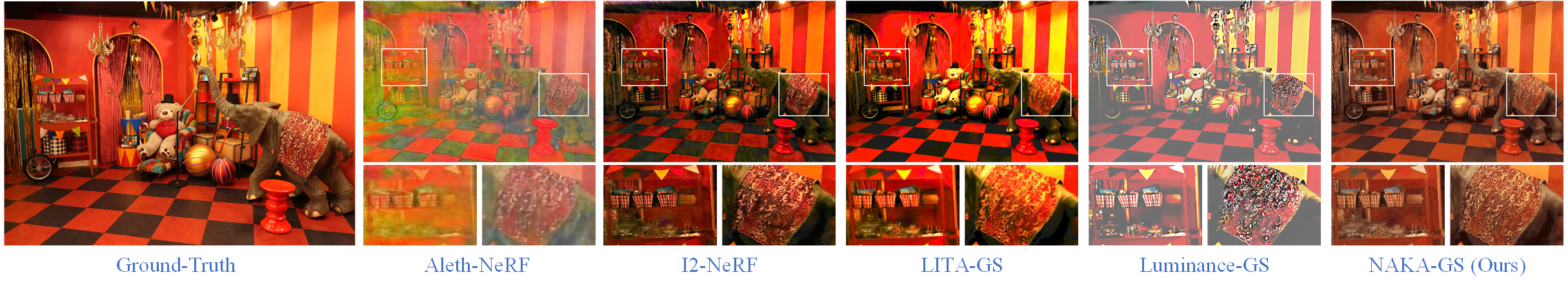}
    \caption{Representative qualitative comparison on MilkCookie, Laboratory, and Popcorn of RealX3D low-light scenes. Full-scene results are provided in the supplementary material.}
    \label{fig:qualitative_comparison}
\end{figure*}

Figure~\ref{fig:qualitative_comparison} presents representative comparisons under severe low-light degradation. Existing methods frequently exhibit green or yellow color casts, clipped highlights, washed-out regions, and blurred object boundaries. These artifacts are particularly evident around thin structures, highly saturated surfaces, and transitions between illuminated and dark regions. In contrast, Naka-GS produces tones that more closely match the ground truth, suppresses chromatic deviations, and retains clearer local contours and textures. The qualitative results are consistent with the improvements in PSNR, SSIM, and LPIPS, supporting the effectiveness of the proposed chromatic correction, frequency-decoupled reconstruction, and Point-Prior compaction. Comparisons on the \textit{BlueHawaii} and remaining  five scene-disjoint test scenes are provided in the supplementary material.

% Replace the following file names if your scene-wise visualization files use different names.
% The extension is intentionally omitted so LaTeX can resolve .pdf, .png, or .jpg automatically.

\subsection{Ablation Study}

\begin{table}[t!]
\centering
\caption{Stage-wise ablation on the eight scene-disjoint RealX3D test scenes. Naka-only denotes the fixed Naka--Rushton pre-enhancement without learned correction. Pts. is the number of initialization points after point preprocessing, and Avg.~T. is the average 3DGS optimization time per scene.}
\label{tab:stage_ablation}
\small
\setlength{\tabcolsep}{2pt}
\begin{tabular}{@{}lccccc@{}}
\toprule
\textbf{Variant} & \textbf{PSNR}$\uparrow$ & \textbf{SSIM}$\uparrow$ & \textbf{LPIPS}$\downarrow$ & \textbf{Pts.} & \textbf{Avg. T.} \\
\midrule
Raw w/o PPM & 6.601 & 0.0637 & 0.8400 & 1.00M& 387.63 \\
Naka-only w/o PPM & 12.966 & 0.5376 & 0.3594 & 1.00M & 383.56 \\
Naka-only + PPM & 13.008 & 0.5615 & 0.3463 & 61.33K & 251.03 \\
Naka-GCC w/o PPM & 20.557 & 0.7227 & \best{0.3300} & 1.00M & 403.86 \\
Naka-GCC + PPM & \best{20.735} & \best{0.7408} & 0.3340 & 59.52k & 251.53 \\
\bottomrule
\end{tabular}
\end{table}

\begin{table}[t!]
\centering
\caption{Image-space ablation of the Naka-GCC correction module on the eight scene-disjoint RealX3D test scenes. HF denotes high-frequency residual preservation, and $\mathcal{L}_{\mathrm{gray}}$ denotes gray-edge supervision.}
\label{tab:naka_internal_ablation}
\small
\setlength{\tabcolsep}{4pt}
\begin{tabular}{@{}lccc@{}}
\toprule
\textbf{Variant} & \textbf{PSNR}$\uparrow$ & \textbf{SSIM}$\uparrow$ & \textbf{LPIPS}$\downarrow$ \\
\midrule
Full & \best{21.269} & 0.7058 & \best{0.3218} \\
w/o $\mathbf{M}_{\text{add}}$ & 18.566 & 0.5725 & 0.3917 \\
w/o $\mathbf{M}_{\text{mul}}$ & 21.169 & 0.6987 & 0.3330 \\
w/o HF & 19.851 & 0.6915 & 0.3294 \\
w/o $\mathcal{L}_{\text{gray}}$ & 20.557 & \best{0.7071} & 0.3221 \\
\bottomrule
\end{tabular}
\end{table}

\paragraph{Contributions of Naka pre-enhancement, Naka-GCC, and PPM.}
Table~\ref{tab:stage_ablation} isolates the contributions of the three stages. Fixed Naka pre-enhancement raises PSNR from 6.601 to 12.966~dB and SSIM from 0.0637 to 0.5376, confirming that the analytical prior recovers low-light visibility. Under the same dense initialization, Naka-GCC further reaches 20.557~dB/0.7227/0.3300, demonstrating the necessity of learned chromatic and structural correction. PPM then reduces the point prior from 1.00M to 59.52K and the pipeline time from 403.86 to 251.53~s, while maintaining comparable reconstruction quality. Thus, Naka-GCC primarily improves fidelity, whereas PPM improves compactness and efficiency.

\paragraph{Internal design of Naka-GCC.}
Table~\ref{tab:naka_internal_ablation} evaluates the internal design of Naka-GCC in image space. Removing $M_{\mathrm{add}}$ causes the largest degradation, identifying additive RGB correction as the main mechanism for compensating chromatic errors. Removing HF also consistently degrades all metrics, confirming the benefit of a frequency-decoupled strategy. Removing $M_{\mathrm{mul}}$ leads to modest but consistent drops in PSNR and SSIM and a higher LPIPS, showing that multiplicative correction provides complementary luminance refinement. Removing $\mathcal{L}_{\mathrm{gray}}$ lowers PSNR and slightly worsens LPIPS, although SSIM increases marginally, indicating a mild distortion--structure trade-off. Overall, additive correction and high-frequency preservation are the dominant components.
\section{Conclusion}
\label{sec:conclusion}

We presented \textbf{NAKA-GS}, a dedicated framework for low-light 3D Gaussian Splatting that integrates biologically motivated intensity adaptation, learnable chromatic correction, and efficient Gaussian initialization within a coherent reconstruction workflow. At its core, NAKA-GS transforms the fixed Naka response into a learnable, frequency-aware correction process, recovering visibility while maintaining color fidelity and fine image content under severe illumination degradation. The resulting formulation remains fully compatible with the standard 3DGS representation and rendering objective. Experiments on eight scene-disjoint RealX3D scenes show that NAKA-GS consistently achieves the best reconstruction performance among the evaluated low-light NeRF and 3DGS methods, while substantially reducing the computational cost of Gaussian optimization. These results demonstrate the effectiveness of NAKA-GS as an accurate and efficient solution for low-light novel-view synthesis.

\bibliography{main}

% The AAAI-27 portal may request the reproducibility checklist separately.
% Uncomment only if the official track instructions require it in the PDF.
% \input{ReproducibilityChecklist.tex}

\end{document}